\documentclass[]{fairmeta}

\usepackage{graphicx} 
\usepackage{wrapfig}
\usepackage{booktabs}
\usepackage{multirow}
\usepackage{makecell}
\usepackage{pifont} 
\usepackage{xcolor} 
\usepackage{listings}
\usepackage{listingsutf8}

\usepackage[utf8]{inputenc}
\usepackage[table]{xcolor}      
\definecolor{lightblue}{RGB}{230,240,255} 
\usepackage{caption}            

\usepackage{graphicx} 
\usepackage{wrapfig}
\usepackage{booktabs}
\usepackage{multirow}
\usepackage{makecell}
\usepackage{pifont}
\usepackage{colortbl}
\usepackage{xcolor}
\usepackage{array}
\usepackage{caption}
\usepackage{enumitem}

\definecolor{headerblue}{HTML}{EBF3FF}
\definecolor{textred}{HTML}{980000}
\definecolor{textgreen}{HTML}{274E13}
\definecolor{mygray}{gray}{0.5}

\definecolor{lightblue}{RGB}{220,230,241}
\definecolor{graytext}{gray}{0.4}

\definecolor{darkblue}{rgb}{0.0,0.0,0.7}
\definecolor{darkred}{rgb}{0.7,0.0,0.0}

\title{Action100M: A Large-scale Video Action Dataset}

\author[1,2]{Delong Chen}
\author[1,3]{Tejaswi~Kasarla}
\author[1]{Yejin~Bang}
\author[1,4]{Mustafa~Shukor}
\author[1,4]{Willy~Chung}
\author[1]{Jade~Yu}
\author[1]{Allen~Bolourchi}
\author[1]{Théo Moutakanni}
\author[1,2]{Pascale~Fung}

\affiliation[1]{Meta FAIR}
\affiliation[2]{HKUST}
\affiliation[3]{University of Amsterdam}
\affiliation[4]{Sorbonne Université}

\abstract{
Inferring physical actions from visual observations is a fundamental capability for advancing machine intelligence in the physical world. Achieving this requires large-scale, open-vocabulary video action datasets that span broad domains. 
We introduce \textsc{Action100M}, a large-scale dataset constructed from 1.2M Internet instructional videos (14.6 years of duration), yielding $O(100 \text{ million})$ temporally localized segments with open-vocabulary action supervision and rich captions. \textsc{Action100M} is generated by a fully automated pipeline that (i) performs hierarchical temporal segmentation using \texttt{V-JEPA 2} embeddings, (ii) produces multi-level frame and segment captions organized as a \textsc{Tree-of-Captions}, and (iii) aggregates evidence with a reasoning model (\texttt{GPT-OSS-120B}) under a multi-round \textsc{Self-Refine} procedure to output structured annotations (brief/detailed action, actor, brief/detailed caption).
Training \texttt{VL-JEPA} on \textsc{Action100M} demonstrates consistent data-scaling improvements and strong zero-shot performance across diverse action recognition benchmarks, establishing Action100M as a new foundation for scalable research in video understanding and world modeling.}

\correspondence{\email{delong.chen@connect.ust.hk}, \email{theomoutakanni@meta.com} }

\metadata[Dataset]{\url{https://github.com/facebookresearch/Action100M}}

\begin{document}

\maketitle

\section{Introduction}
\label{sec:intro}

Making machine intelligence useful in the physical world requires AI models that not only understand world states (\textit{e.g.,} objects and their attributes), but also recognize \textbf{physical actions} that interact with the world and induce state transitions. Powered by supervision from large datasets~\citep{laurenccon2023obelics, awadalla2024mint, shukor2025scaling, schuhmann2022laion}, world state perception in frontier models has advanced rapidly~\citep{alayrac2022flamingo, chen2022pali,Bai2023QwenVLAV, liu2024improved, dai2023instructblip, marafioti2025smolvlm}. In contrast, the capability of understanding \textit{actions} is comparably limited, largely due to the absence of robust data foundations. Existing video action datasets are mainly separately developed for individual narrow domains (\textit{e.g.,} cooking, toy assembly), and remain in limited scale (\textit{e.g.,} less than 1 million action instances). 

Advancement in data foundation will enable the development of \textbf{open-domain} and \textbf{open-vocabulary} video action recognizers. These models will support embodied learning, facilitate wearable assistive applications, and advance physical world modeling~\citep{chen2025planning, chen2025worldprediction, fung2025embodied, terver2025drives, genie3, worldlabs, russell2025gaia, agarwal2025cosmos, xiang2024pandora}. Achieving this necessiates training data that is sufficiently \textit{large scale}, maintains \textit{high quality}, and spans \textit{broad diversity}. 

To address these challenges, we introduce \textsc{Action100M}, a large-scale video action dataset containing $O(100 \text{ million})$ action instances, annotated via a fully automated pipeline. The construction process begins with online instructional videos~\citep{miech2019howto100m}, which capture people interacting with the physical world in diverse activities. Dense action labels are produced using a family of frontier open-source models, including \texttt{V-JEPA 2}~\citep{assran2025v}, \texttt{PerceptionLM}~\citep{cho2025perceptionlm}, \texttt{Llama-3.2-Vision}, and \texttt{GPT-OSS}~\citep{agarwal2025gpt}. The \textsc{Tree-of-Captions} ~\citep{chen2024makes} and \textsc{Self-Refine} mechanisms~\citep{chen2025planning} are leveraged to reduce hallucinations and ensure annotation quality. This pipeline yields a hierarchy of temporal segments annotated with structured fields, including brief/detailed action descriptions, the actor, and brief/detailed video captions. 

\looseness=-1
The annotation pipeline was executed on 1.2 million YouTube videos (14.6 years of duration), spending approximately 1.3 million V100 GPU hours for segmentation and captioning, and 0.3 million H100/H200 GPU hours for LLM aggregation. It derives 147 million segment-level annotations, totaling 21.3 billion English words.

To demonstrate the utility of \textsc{Action100M}, we train \texttt{VL-JEPA} models \citep{chen2025vl} on the dataset. \texttt{VL-JEPA} is a vision-language model that extracts visual embeddings with \texttt{V-JEPA 2} \citep{assran2025v} encoder and predicts target embedding generated by a text encoder. The model uses InfoNCE loss to learn an aligned embedding space that allows CLIP-style open-vocabulary classification \citep{radford2021learning}. 
Evaluation is conducted on eight downstream video action benchmarks, including Something-something-v2~\citep{goyal2017something}, EPIC-KITCHENS-100~\citep{damen2020epic}, EgoExo4D Keysteps~\citep{grauman2024ego}, Kinetics-400~\citep{kay2017kinetics}, COIN~\citep{tang2019coin}, and CrossTask~\citep{zhukov2019cross}. These benchmarks evaluate actions spanning diverse domains and different levels of abstraction, from low-level fine-grained ones like \textit{``turn on light''}, \textit{``cut tomato''}, to high-level procedural ones like \textit{``make Irish coffee''}, \textit{``assemble computer''}, etc. 

\begin{wrapfigure}{r}{0.38\textwidth}
    \centering
    \vspace{-10pt}
    \includegraphics[width=\linewidth]{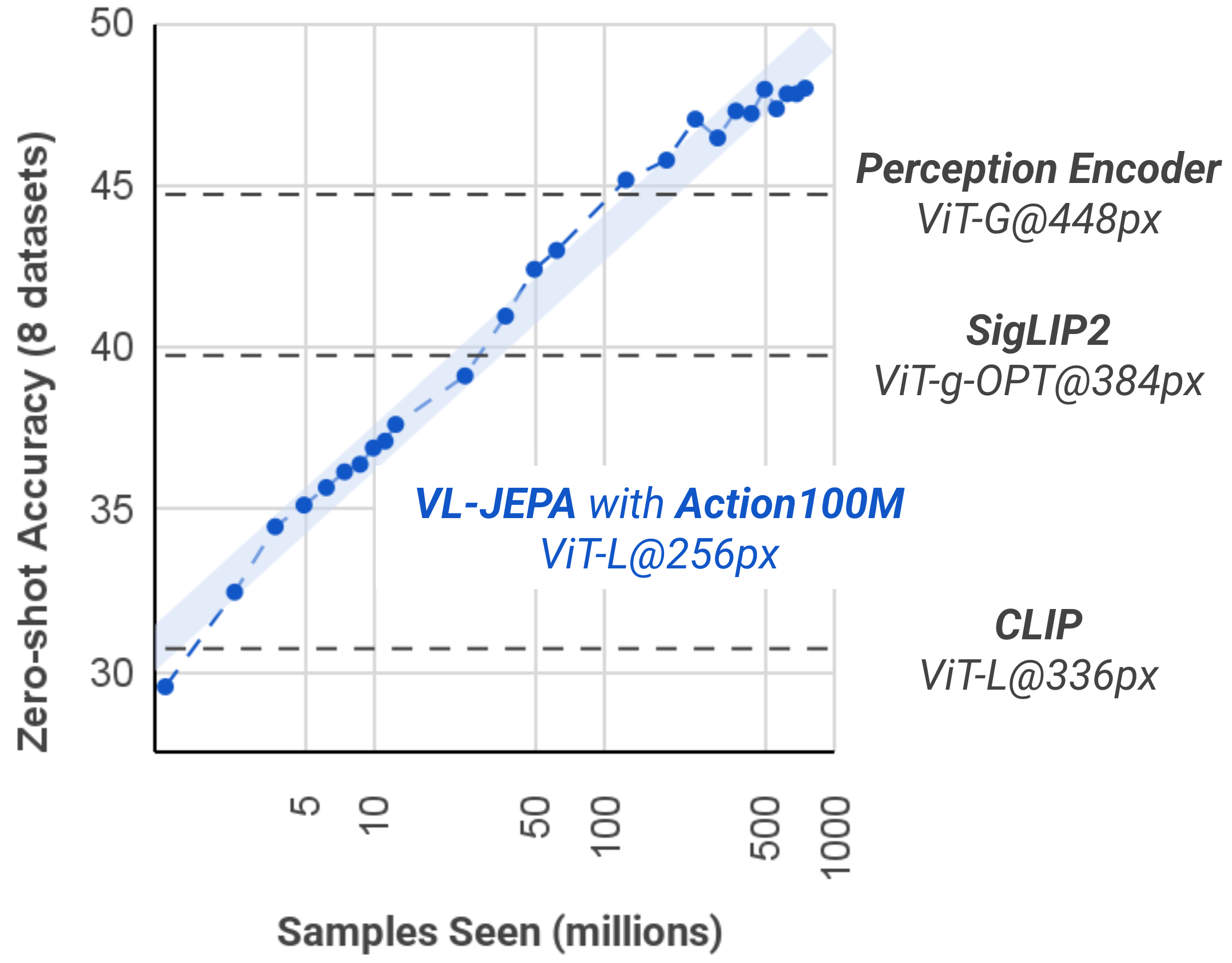}
    \caption{Scaling on \textsc{Action100M} improves zero-shot action recognition consistently.}
    \vspace{-10pt}
    \label{fig:scaling_curve}
\end{wrapfigure}

Fig.~\ref{fig:scaling_curve} demonstrates the \textbf{effectiveness of scaling} for zero-shot action recognition: \texttt{VL-JEPA} performance improves consistently as the amount of \textsc{Action100M} training data increases (x-axis shows effective batch size $\times$ number of iterations). Our \texttt{VL-JEPA} (ViT-L at 8 frames per input in 256px) outperform CLIP \citep{radford2021learning} , SigLIP2 \citep{tschannen2025siglip}, and Perception Encoder \citep{bolya2025perception} which have seen much more samples (13B, 40B, and 86B, respectively, while VL-JEPA only 3B) and use larger backbone and higher input resolution. At the same time, the caption annotations in \textsc{Action100M} allows \texttt{VL-JEPA}  to establish robust vision-language alignment that allows effective zero-shot text-to-video retrieval. Compared to CLIP, SigLIP2, and Perception Encoder, \texttt{VL-JEPA} achieves higher average retrieval recall@1 on eight benchmarks: MSR-VTT \citep{xu2016msr}, ActivityNet \citep{caba2015activitynet}, DiDeMo \citep{anne2017localizing}, MSVD \citep{chen2011collecting}, YouCook2 \citep{zhou2018towardsyoucook}, PVDBench \citep{bolya2025perception}, Dream-1K \citep{wang2024tarsier}, and VDC-1K \citep{chai2024auroracap}.

\section{Related Works}
\label{sec:related_work}

\begin{table*}[h]
\centering
\caption{\textbf{Summary of major video caption and action recognition datasets.} This table summarizes key statistics for prominent video datasets, including total duration, number of videos and clips, average text length, and annotation sources. The top table lists large-scale video caption datasets and the bottom table presents action recognition and instructional video datasets, including both manual and automated annotations. \textsc{Action100M} is distinguished by its unprecedented scale (100M clips), rich hierarchical annotations (brief and detailed actions and captions), and broad coverage of real-world activities.}
\label{tab:video_datasets_combined}

\begin{subtable}{\textwidth}
\centering
\caption{Action Recognition and Instructional Video Datasets}
\resizebox{\textwidth}{!}{%
\begin{tabular}{lccccl}
\toprule
\textbf{Dataset} & \textbf{Duration} & \textbf{\#Videos} & \textbf{\#Clips} & \textbf{Avg Text Length}& \textbf{Annotation} \\
\midrule
COIN~\citep{tang2019coin}                  & 476 hours   & 11.8K  & 46.3K  & 4.8 & Manual                     \\
YouCook2~\citep{zhou2018towardsyoucook}    & 176 hours   & 2K     & 14K    & 8.8 & Manual                     \\
THUMOS14~\citep{idrees2017thumos}         & 30 hours & 2,584 & 20,108 && Manual                         \\
ActivityNet Captions~\citep{caba2015activitynet}  & 849 hours   & 20K    & 100K   & 13.5 & Manual                     \\
FineAction~\citep{liu2022fineaction}       & 705 hours   & 17K    & 103K   & -- & --                         \\
EGTEA~\citep{sudhakaran2021learning}       & 28 hours & 86 & 10,325 & & Manual                 \\
50Salads~\citep{50Salads}                  & 4 hours     & --     & --     &-- & --                         \\
Breakfast~\citep{breakfast}                & 77 hours    & --     & 11,267 & -- & Manual                     \\
Assembly101~\citep{sener2022assembly101}   & 513 hours   & 4321   & 1M     &-- & Manual                     \\
EgoProceL~\citep{bansal2022my}             & 62 hours    & 329    & --     &-- & Manual                     \\
Ego4D-Goal-step~\citep{song2023ego4d}      & 368 hours   & 851    & 48K   & -- & Manual                     \\
\midrule
Action100M Brief Action                             & 14.6 years  & 1.2M   & 147M   & 18.4 & PLM-3B, Llama-3.2-Vision-11B, GPT-OSS-120B       \\
Action100M Detailed Action                             & 14.6 years  & 1.2M   & 147M   & 150.2  & PLM-3B, Llama-3.2-Vision-11B, GPT-OSS-120B       \\
\bottomrule
\end{tabular}}
\end{subtable}

\vspace{1em}

\begin{subtable}{\textwidth}
\centering
\caption{Caption Datasets}
\resizebox{\textwidth}{!}{%
\begin{tabular}{lccccl}
\toprule
\textbf{Dataset} & \textbf{Duration} & \textbf{\#Videos} & \textbf{\#Clips} & \textbf{Avg Text Len} & \textbf{Annotation} \\
\midrule
YT-Temporal-180M~\citep{zellers2021merlot}      & --         & 6M     & 180M   & --    & ASR                        \\
HD-VILA-100M~\citep{xue2022advancing}           & 42.4 years & 3.3M   & 103M   & 32.5  & ASR                        \\
InternVid~\citep{wang2024internvid}             & 86.8 years & 7.1M   & 234M   & 17.6  & Tag2Text, BLIP2            \\
VidChapters-7M~\citep{yang2023vidchapters}      & 35.1 years & 817K   & 6.8M   & --    & ASR, user chapters         \\
Panda-70M~\citep{chen2024panda}                 & 19.0 years & 70.8M  & --     & 13.2  & VideoLlama, MiniGPT-4, etc \\
ShareGPT4Video~\citep{chen2024sharegpt4video}   & 291 hours  & 40K    & --     & 273.3 & GPT-4V                     \\
OpenVid-1M~\citep{nan2025openvidm}              & 0.23 years & 1M     & --     & 126.5 & LLaVA-v1.6-34B             \\
MiraData~\citep{ju2024miradata}                 & 1.8 years  & 330K   & --     & 318   & GPT-4V                     \\
VidGen-1M~\citep{tan2024vidgen}                 & 0.34 years & 1M     & --     & 89.3  & VILA, Llama-3.1            \\
Koala-36M~\citep{wang2025koala}                 & 19.6 years & 36M    & --     & 202.1 & GPT-4V                     \\
EgoVid-5M~\citep{wang2024egovid}                & 0.63 years & 5M     & --     & --    & LLaVA-NeXTVideo, Qwen2     \\
OpenHumanVid~\citep{li2025openhumanvid}         & 1.9 years  & 13.2M  & --     & --    & Llama 3.1, BLIP2           \\
VideoUFO~\citep{wang2025videoufo}               & 0.44 years & 568K   & 1M     & 155.5 & Qwen2-VL-7B                \\
UltraVideo~\citep{xue2025ultravideo}            & 62 hours   & 5K     & 42K    & 824.2 & Qwen2.5-VL-72B             \\
PE Video Dataset~\citep{cho2025perceptionlm}      & 0.49 years         & 1M    & 120K     &  111.7    & PLM, Llama-3.3-70B, human refined                         \\
PLM-Video-Auto~\citep{cho2025perceptionlm}      & 6.06 years         & 6.4M    & --     & --   & Llama-3.3-70B, LLama-3-405B \\ 
\midrule
Action100M Brief Caption                                & 14.6 years  & 1.2M   & 147M    & 106.8 & PLM-3B, Llama-3.2-Vision-11B, GPT-OSS-120B       \\
Action100M Detailed Caption                             & 14.6 years  & 1.2M   & 147M    & 540.0 & PLM-3B, Llama-3.2-Vision-11B, GPT-OSS-120B       \\
\bottomrule
\end{tabular}}
\end{subtable}

\end{table*}

We present a comparison of \textsc{Action100M} with existing action and caption datasets in Table~\ref{tab:video_datasets_combined}, and discuss the detailed related work below. We organize prior efforts into two main categories: video action datasets, which focus on annotated action segments, and video captioning datasets, which provide descriptive text for video content.

\textbf{Video Action Datasets}. 
Standard video action datasets, such as COIN~\citep{tang2019coin}, CrossTask~\citep{zhukov2019cross}, and YouCook2~\citep{zhou2018towardsyoucook}, have played a pivotal role in advancing action detection in untrimmed videos. These datasets typically segment long activities into semantically meaningful steps, each paired with textual descriptions, and are primarily composed of instructional videos covering procedural tasks.
The collection methodologies for video datasets vary significantly:
\begin{itemize}
    \item Participant-recorded datasets (e.g., Breakfast~\citep{breakfast}, Assembly101~\citep{sener2022assembly101}) provide dense, hierarchical annotations in controlled environments with high-quality labels through manual data collection.
    \item Internet-mined datasets (e.g., CrossTask, COIN) leverage external taxonomies to achieve broader coverage of activities. By mining videos from online sources, these datasets capture a wider range of human actions.
    \item Egocentric datasets (e.g., EgoProceL~\citep{bansal2022my}, Ego4D Goal-Step~\citep{song2023ego4d}) introduce data-driven, hierarchical taxonomies and dense step annotations, particularly in domains such as cooking. These datasets support research on goal inference and long-term temporal modeling by capturing first-person perspectives and fine-grained activity sequences.
\end{itemize}

Despite these advances, existing datasets are often constrained by manual annotation bottlenecks, limited domain coverage, or lack of scale.

While datasets such as COIN~\citep{tang2019coin} and YouCook2~\citep{zhou2018towardsyoucook} provide valuable annotated instructional videos, they are constrained by manual annotation and limited activity coverage. Internet-mined datasets expand diversity but often lack fine-grained, hierarchical labels. Egocentric datasets introduce new perspectives but remain domain-specific. \textsc{Action100M} marks a step forward in this landscape.

In contrast, \textsc{Action100M} leverages large-scale online instructional videos~\citep{miech2019howto100m} to deliver over 100 million action instances with multi-level, open-vocabulary annotations. This enables robust research in open-domain action recognition and world modeling, as summarized in Table~\ref{tab:video_datasets_combined}. By addressing annotation bottlenecks and supporting fine-grained temporal reasoning, \textsc{Action100M} provides the way for future advances in video understanding.

\textbf{Video Caption Datasets}. 
A number of large-scale video–language datasets have been proposed in recent years. InternVid~\citep{wang2024internvid}, HD-VILA-100M~\citep{xue2022advancing}, and VidChapters-7M~\citep{yang2023vidchapters} collect millions of videos or hundreds of millions of clips with associated textual descriptions. However, the majority of these captions are obtained from noisy sources such as automatic speech recognition (ASR) transcripts or video metadata. As a result, the captions are typically short, generic, and only weakly aligned with the underlying physical actions, limiting their usefulness for learning fine-grained action representations or detailed world models.

To address the limitations of ASR-only captions, recent instruction-style datasets leverage powerful vision–language models to produce richer descriptions. OpenVid-1M~\citep{nan2025openvidm}, UltraVideo~\citep{xue2025ultravideo}, and VideoUFO~\citep{wang2025videoufo} generate longer, more detailed, and instruction-like captions that better capture object interactions and step-by-step activities. Nonetheless, these datasets are typically built on a relatively small number of source videos compared to the largest ASR-based corpora, which constrains their coverage of diverse environments and long-horizon activities. Koala-36M~\citep{wang2025koala} further scales LLM-captioned video data, but is primarily oriented toward video generation rather than structured action understanding.

For video understanding and perception-centric applications, PE Video and PLM-Video-Auto~\citep{cho2025perceptionlm} combine millions of videos with LLM-generated captions. While they improve caption quality over pure ASR pipelines, they generally assign a single caption per segment and do not explicitly encode temporal hierarchies within videos. Consequently, they provide limited supervision for modeling multi-scale structure in activities (e.g., steps, sub-tasks, and overarching tasks), which is crucial for learning world models that reason over extended sequences of actions and their context.

Moreover, many existing video captioning datasets rely heavily on manual or semi-manual annotation, which is expensive and restricts scale, or they prioritize scale at the cost of fine-grained, action-centric detail and temporal structure. In contrast, \textsc{Action100M} is constructed to combine the scale of ASR-driven corpora with the descriptiveness of recent LLM-captioned datasets. It introduces a Tree-of-Captions structure derived from hierarchical temporal segmentation, providing multiple levels of captions that capture both fine-grained actions and broader contextual narratives. This design preserves temporal hierarchy in a dense-captioning-like manner, making \textsc{Action100M} particularly suited for training large-scale world models and action-centric video understanding systems.

\section{Action100M Data Pipeline}
\label{sec:action100m_pipeline}

\begin{figure}
    \centering
    \includegraphics[width=1\linewidth]{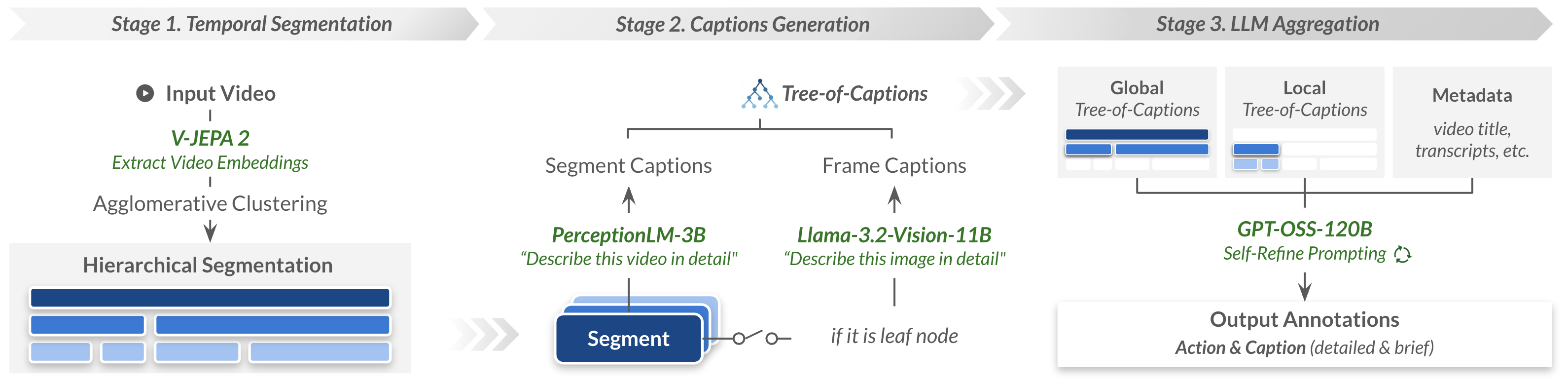}
    \caption{\textbf{\textsc{Action100M} Data Pipeline.} Our pipeline first applies hierarchical temporal segmentation to decompose the video into semantically coherent segments at multiple temporal scales. For each segment, we generate video caption and frame captions, capturing both temporal and spatial information. Next, we prompt LLM to aggregate the captions, extracting final annotations.}
    \label{fig:data-pipeline}
\end{figure}

\begin{figure}
    \centering
    \includegraphics[width=1\linewidth]{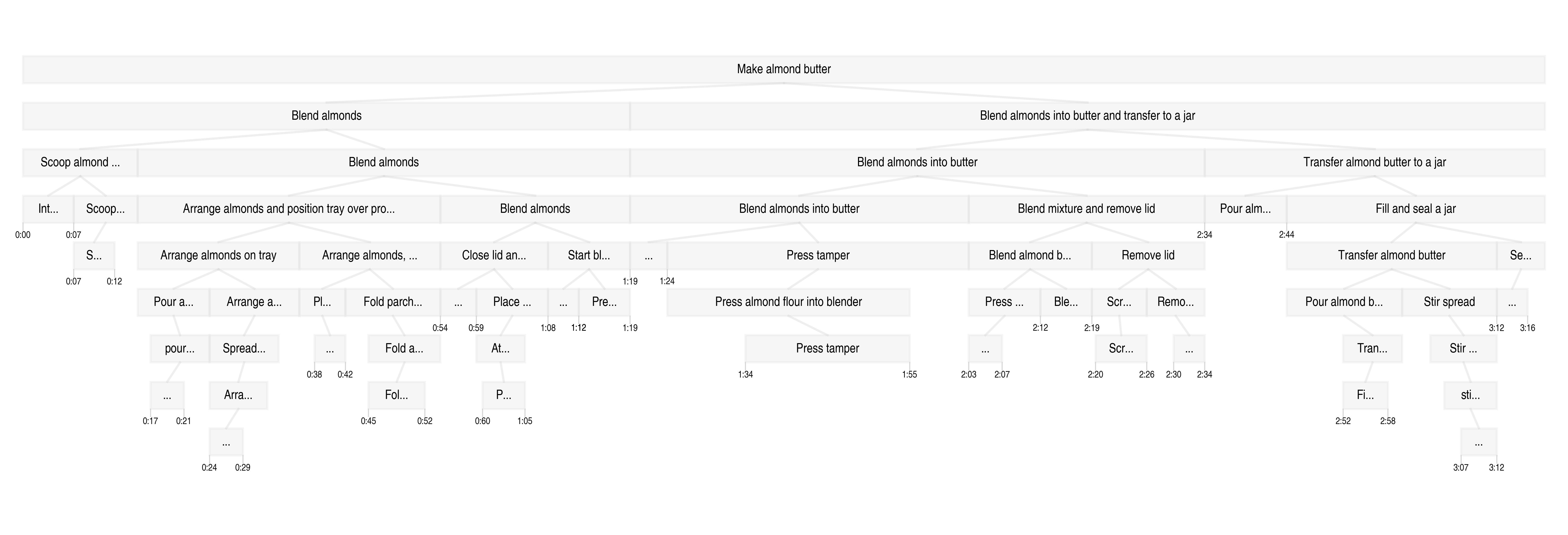}
    \vspace{-30pt}
    \caption{\textbf{Example of hierarchical structure in Action100M annotations} (with brief action description labels shown). Source video: \href{https://www.youtube.com/watch?v=NYRlBWgLbKU}{url}. \textbf{Brief caption} of the entire video: \textit{A woman roasts almonds, blends them into butter, and pours the butter into a jar.} \textbf{Detailed caption}: \textit{The video opens with the presenter in a bright kitchen speaking to the camera. She spreads raw almonds on a parchment‑lined tray, places the tray in a pre‑heated 350 °F oven, and after roasting lets the nuts cool to room temperature. She then transfers the almonds to a Vitamix blender, removes the lid, inserts a tamper, and sets the machine on high. While the blender runs she presses the almonds down with the tamper, first creating a fine flour and then a thick creamy butter within about one minute. She pours the almond butter into a clear storage jar, scoops it with a large wooden spoon and stirs it to smooth the surface, then concludes the segment with a brief thank‑you.}}
    \label{fig:action100m_tree_NYRlBWgLbKU}
\end{figure}

This section describes an automated, scalable pipeline for constructing \textsc{Action100M}. The pipeline extends and improves the data generation procedure introduced in Vision Language World Model~\citep{chen2025planning}. Fig.~\ref{fig:data-pipeline} provides an overview of the pipeline, and Fig.~\ref{fig:action100m_tree_NYRlBWgLbKU} visualizes an example of annotation. The pipeline has three stages: 

\begin{enumerate}
    \item We decompose each video into a hierarchy of temporally coherent segments, which captures both short, fine-grained motions and long, procedural steps, enabling action supervision at multiple levels of abstraction. 

    \item For every segment, we generate complementary frame- and video-level captions and organize them into a \textsc{Tree-of-Captions}. Captioning is a task for which modern VLMs are extensively trained, and even medium-sized models can produce reliable results. 

    \item We prompt an LLM to aggregate information from the \textsc{Tree-of-Captions} and to extract structured annotations. Assembling evidence across multiple caption levels provably reduces hallucinations~\citep{chen2024makes}. 
\end{enumerate}

Overall, rather than applying heavy VLMs directly to entire videos, our pipeline first converts videos into hierarchical, text-based representations (\textsc{Tree-of-Captions}, inspired by Pyramid-of-Captions~\citep{chen2024makes}) and only leverages strong reasoning models in the final stage, which is purely text-based. This design allow us to obtain reliable annotations while keeping the overall computation cost manageable. In the following, we describe each stage in detail.

\textbf{Stage 1. Temporal Segmentation.}  Each video is first transformed into a temporally dense sequence of visual embeddings using the \texttt{V-JEPA~2} encoder. Frames are uniformly sampled at one out of every four raw frames to approximate the temporal resolution used during V-JEPA 2 pretraining. The video is then divided into overlapping temporal windows, each containing 64 sampled frames, with an eight-frame stride between consecutive windows. Each window is independently processed by the V-JEPA 2 ViT-g-384\footnote{\url{https://huggingface.co/facebook/vjepa2-vitg-fpc64-384}} encoder, which outputs a sequence of spatial-temporal visual tokens. We perform spatial average pooling, resulting in a per-frame feature of dimensionality equal to the encoder’s hidden size. Because adjacent windows overlap, multiple representations are produced for shared frames. These are accumulated and averaged to form a single, temporally consistent embedding per frame across the entire video.

To capture actions across multiple temporal scales, we apply \textit{hierarchical agglomerative clustering} to the sequence of frame-level representations\footnote{\url{https://scikit-learn.org/stable/modules/generated/sklearn.cluster.AgglomerativeClustering.html}}. A local temporal connectivity constraint links each frame only to its immediate neighbors, ensuring that merges occur only between contiguous time spans. Clustering proceeds bottom-up using Ward linkage, which minimizes the variance within each segment at every merge step. Starting from individual frames or short windows, adjacent segments are iteratively merged to minimize intra-cluster variance. This process produces a hierarchical tree decomposition of the video, where each node corresponds to a contiguous, semantically coherent temporal segment. Lower levels of the hierarchy correspond to fine-grained atomic motions, while higher levels capture coarser activities. We retain only nodes whose duration is larger than 0.5 seconds, ensuring that each segment is semantically meaningful.

\textbf{Stage 2. Caption Generation.}  
After hierarchical segmentation, we annotate each node in the video tree with captions, ensuring that both local fine-grained and global contextual information are captured within the same representation. The captioning process operates in two complementary modes: mid-frame captioning for fine-grained spatial details and video-segment captioning for temporal dynamics. 

For every leaf node (representing the smallest contiguous action segment), we extract a key frame at the midpoint of its temporal span. These mid-frames are processed using \texttt{Llama-3.2-Vision-11B}\footnote{\url{https://huggingface.co/meta-llama/Llama-3.2-11B-Vision}}, prompted with \textit{``Describe this image in detail.''} to generate frame captions. For higher-level nodes representing longer temporal spans, we apply \texttt{Perception-LM-3B}\footnote{\url{https://huggingface.co/facebook/Perception-LM-3B}}. Each segment is sampled into 32 evenly spaced frames between its start and end times at $320^2$ resolution, and the model is prompted with \textit{``Describe this video in detail.''} to generate segment-level captions. Both models are configured with a generation limit of 1024 tokens and can be run on a single NVIDIA V100 32GB GPU.

\textbf{Stage 3. LLM Aggregation.}  
For each node in the \textsc{Tree-of-Captions}, we generate structured action annotations by prompting \texttt{GPT-OSS-120B}\footnote{\url{https://huggingface.co/openai/gpt-oss-120b}} to extract five fields of information: brief action description, detailed action description, actor, brief video caption, and detailed video caption. This extraction is performed through three iterative rounds of \textsc{Self-Refine} to improve consistency and quality.

The inputs to the LLM include the current node’s caption, its children’s captions formatted in depth-first order, as well as global context such as root captions (within a limited depth) and video metadata (including title, description, and ASR transcript). Each node in the \textsc{Tree-of-Captions} is processed independently. Nodes with a duration shorter than four seconds are discarded. For the remaining nodes, we query \texttt{GPT-OSS-120B} to infer clean, structured textual representations that unify and denoise information from multiple caption sources. More details about the process and the prompt to the \texttt{GPT-OSS-120B} are in the supplementary.

\section{Dataset Analysis}

\begin{figure}[t]
    \centering
    \includegraphics[width=\linewidth]{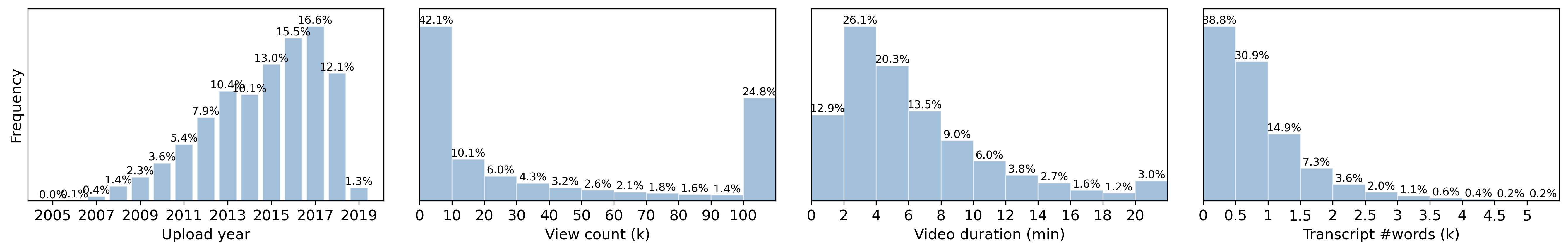}
    \caption{\textbf{Statistics of Action100M source videos and metadata.} Distributions of (left to right) video upload year, view count, video duration, and transcript length, computed over the subset of videos for which metadata is available.}
    \label{fig:metadata_distributions}
\end{figure}

\begin{figure}[t]
    \centering
    \includegraphics[width=1\linewidth]{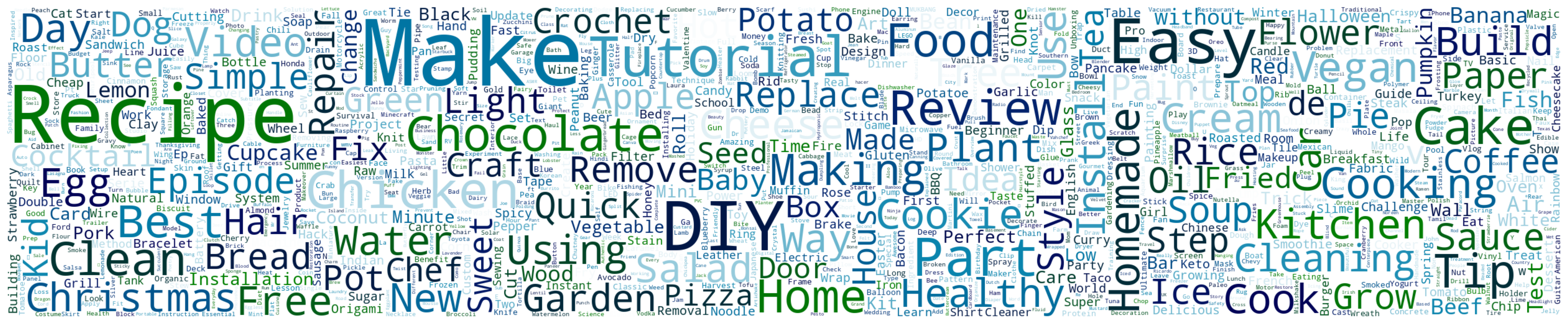}
    \caption{\textbf{Word cloud of video titles in Action100M.} Frequently occurring words reflect the instructional and procedural nature of the dataset, with dominant terms related to cooking, DIY activities, and everyday physical tasks.}
    \label{fig:wordcloud}
\end{figure}

\textbf{Source Videos}. Action100M is built on 1,199,096 face-blurred videos sourced from HowTo100M \citep{miech2019howto100m}, corresponding to a total video duration of approximately 14.6 years. Since the original HowTo100M dataset was released in June 2019~\citep{miech2019howto100m}, many videos have became unavailable since then.  We successfully retrieved ASR transcripts for 72\% of these videos, covering 10.6 years of video content. 
Figure~\ref{fig:metadata_distributions} summarizes the statistics of the available metadata. 
Figure~\ref{fig:wordcloud} visualizes a word cloud derived from video titles. As HowTo100M focuses on instructional content curated from 12 WikiHow categories (e.g., Food \& Entertaining, Home \& Garden, Hobbies \& Crafts, while excluding more abstract categories such as Relationships or Finance), the most frequent keywords strongly reflect procedural and hands-on activities. Common terms include \textit{``make''}, \textit{``recipe''}, \textit{``DIY''}, \textit{``easy''}, \textit{``cake''}, and \textit{``chocolate''}, highlighting the dominance of cooking, home improvement, and everyday skill-oriented videos in the dataset.

\begin{figure}
    \centering
    \includegraphics[width=1\linewidth]{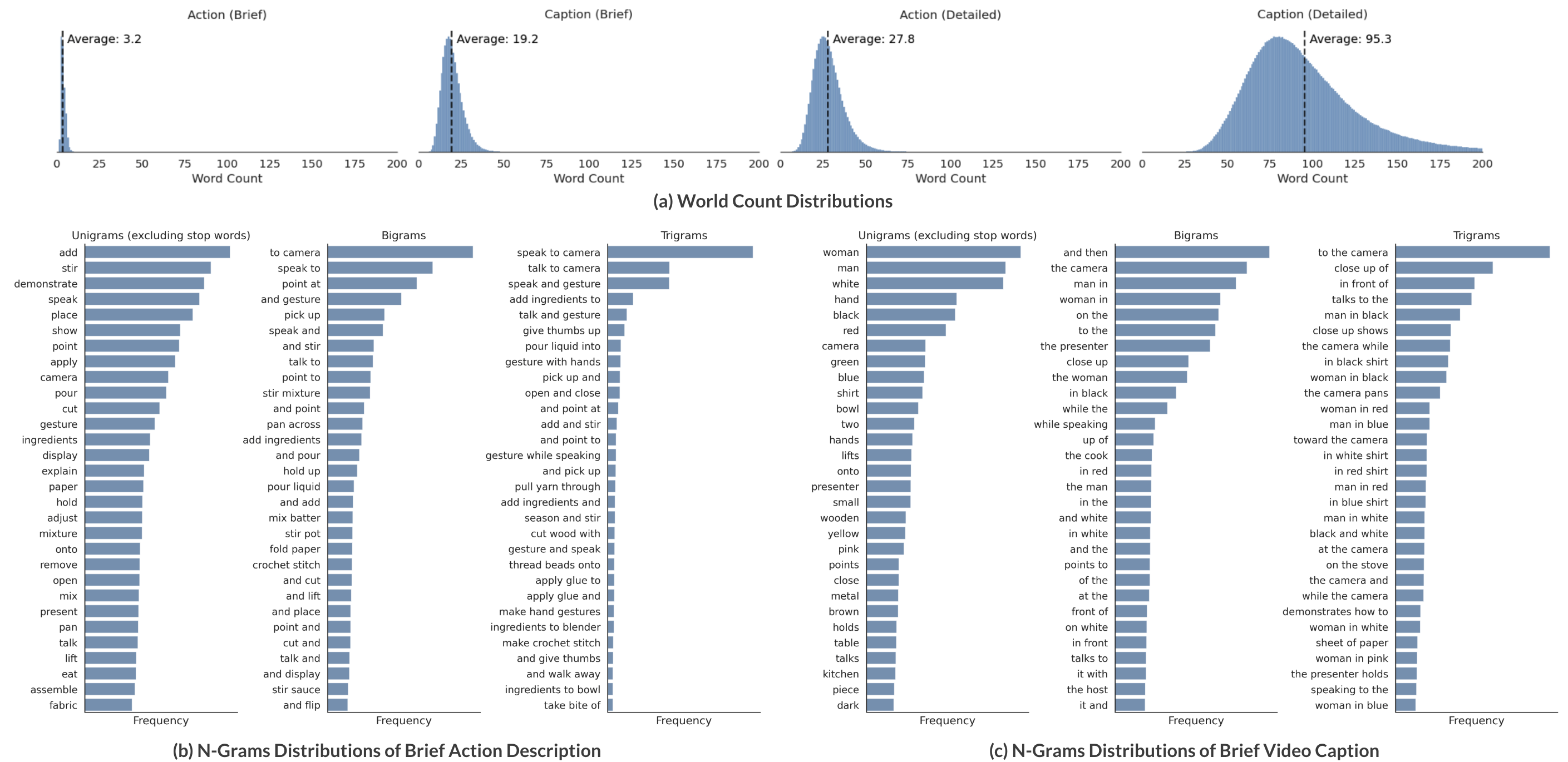}
    \caption{\textbf{Statistics of generated textual annotations in Action100M.} \textbf{(a)} Word-count distributions for four annotation types: brief action, brief caption, detailed action, and detailed caption. Dashed lines indicate the mean length of each annotation type. \textbf{(b-c)} Top unigrams, bigrams, and trigrams (excluding stop words) in brief action descriptions and brief video captions.}

    \label{fig:word_histograms}
\end{figure}

\textbf{Generated Annotations.} Figure~\ref{fig:word_histograms}(a) summarizes the word-count statistics of the four annotation types produced by our pipeline. The average length increases monotonically from brief action (3.2 words), to brief caption (19.2 words), to detailed action (27.8 words), and finally to detailed caption (95.3 words). Across the entire dataset, Action100M contains 147,092,653 annotated video segments (including videos without metadata), corresponding to an estimated total of 0.46B (brief actions) + 2.83B (brief captions) + 3.96B (detailed actions) + 14.02B (detailed captions) = 21.27B words.

Due to the hierarchical temporal segmentation procedure, shorter segments are substantially more common than longer ones. Specifically, 64\% of all segments have durations between 0-3 seconds, followed by 23.8\% in the 3-10 second range, and 10.2\% between 10 seconds and 1 minute. Only about 2\% of segments are longer than one minute. For action annotations, 3.23\% of segments are labeled as \texttt{``N/A''} by \texttt{GPT-OSS-120B}, which typically correspond to non-action content such as video introductions, advertisements, or subscription reminders. Storing all annotations together with metadata and the full Tree-of-Captions structure requires approximately 205~GB of disk space.

\begin{figure}
    \centering
    \includegraphics[width=1\linewidth]{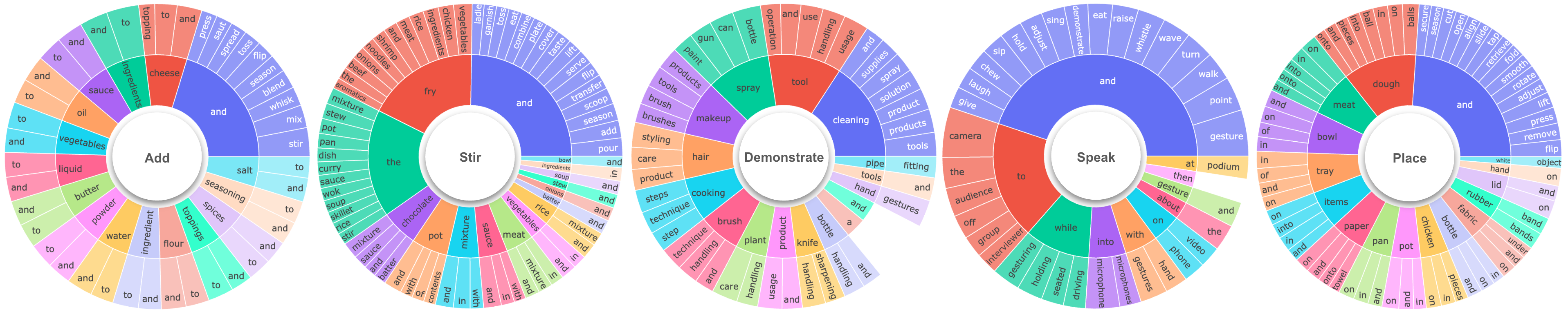}
    \caption{\textbf{Sunburst visualizations of frequent action compositions.} Each sunburst shows the hierarchical co-occurrence structure centered on one of the five most frequent verbs in brief action descriptions.}
    \label{fig:action_sunburst}
\end{figure}

Figures~\ref{fig:word_histograms}(b) and (c) show the N-gram distributions of brief action descriptions and brief video captions, respectively. As expected, action descriptions are dominated by verbs (e.g., \textit{add}, \textit{stir}, \textit{demonstrate}), whereas video captions contain a higher proportion of adjectives and object-centric descriptors. The frequency distributions further reveal a strong imbalance in action concepts, with certain patterns such as \textit{``speak to camera''} occurring disproportionately often. This observation motivates the semantic resampling strategy introduced in \S\ref{sec:semantic resampling} to alleviate long-tail imbalance during training. Figure~\ref{fig:action_sunburst} provides a qualitative view of this structure through sunburst diagrams for the five most frequent words in brief action descriptions.

\section{Experiments}
\label{sec:experiments}

\subsection{Training and Evaluation Setup}

We train \texttt{VL-JEPA} model \citep{chen2025vl}. We perform the query-free pretraining (\textit{i.e.,} \texttt{BASE}) with \textsc{Action100M} data. We refer to \citep{chen2025vl} for details of model architecture. We train \texttt{VL-JEPA} in three stages (Tab.~\ref{tab:training-stages}):

\begin{itemize}
    \item In \textbf{Stage 1}, we perform image pretraining with single frame per input. We use DataComp-1B \citep{gadre2023datacomp}, mixing original caption and model generated caption from \cite{li2024if}, and YFCC-100M \citep{thomee2016yfcc100m} captioned by PaliGemma \citep{beyer2024paligemma} and Florence-2 \citep{xiao2024florence}. 

    \item In \textbf{Stage 2}, we process video data with eight frames per input. We continue from stage 1 checkpoint and use \textsc{Action100M} data (mixing all four field, with detailed action and caption downsampled by a half), and \texttt{PerceptionLM-3B} labelled action description and video caption for each \textsc{Action100M} segment.

    \item In \textbf{Stage 3}, we increase the number of frames per input to 32, and unfreeze the \texttt{V-JEPA 2} encoder. This leads to higher CUDA memory consumption and lower batch size, and we use gradient accumulation of 4 steps and lower learning rate (1e-5 instead of 5e-5).
    
\end{itemize}

\begin{table}[h!]
\caption{Details of \texttt{VL-JEPA}'s three-stage training procedure. Stage 3 uses a gradient accumulation of 4 steps.}
\label{tab:training-stages}
\resizebox{\columnwidth}{!}{%
\begin{tabular}{cccccc}
\toprule
\textbf{Training Stage} & \textbf{Vision encoder} & \multicolumn{1}{l}{\textbf{\#Frames per input}} & \textbf{Training Data} & \textbf{Batch Size} & \textbf{\#Iterations} \\ \midrule
Stage 1 & Frozen & 1 & Image-text data (DataComp-1B, YFCC-100M) & 24,576 & 100k \\
Stage 2 & Frozen & 8 & Action100M & 12,288 & 60k \\
Stage 3 & Unfrozen & 32 & Action100M & 3,072$\times$4 & 10k \\ \bottomrule
\end{tabular}%
}
\end{table}

We benchmark resulting models with existing foundation models, including CLIP \citep{radford2021learning}, SigLIP2 \citep{tschannen2025siglip}, and Perception Encoder \citep{bolya2025perception}. We evaluate on two tasks:

\begin{itemize}
    \item \textbf{Zero-shot action recognition} (top-1 accuracy) on eight benchmarks: Something-something-v2 (SSv2)~\citep{goyal2017something}, EPIC-KITCHENS-100 (EK-100)~\citep{damen2020epic}, EgoExo4D Keysteps~\citep{grauman2024ego}, Kinetics-400~\citep{kay2017kinetics}, COIN~\citep{tang2019coin}, and CrossTask~\citep{zhukov2019cross}. For COIN and Crosstask, we evaluate both segment-level step recognition and global task recognition.

    \item \textbf{Zero-shot text-to-video retrieval} (recall@1) on eight benchmarks: MSR-VTT \citep{xu2016msr}, ActivityNet \citep{caba2015activitynet}, DiDeMo \citep{anne2017localizing}, MSVD \citep{chen2011collecting}, YouCook2 \citep{zhou2018towardsyoucook}, PVD-Bench \citep{bolya2025perception}, Dream-1K \citep{wang2024tarsier}, and VDC-1K \citep{chai2024auroracap}.
\end{itemize}

\subsection{Main Results}

\begin{table*}[]
\caption{\textbf{Main results}. We evaluate zero-shot performance on eight action recognition dataset. and eight video retrieval datasets.
}
\label{tab:cls_ret}
\resizebox{\linewidth}{!}{%
\begin{tabular}{ccrrrc|c|cccccccc|c|cccccccc}
\multicolumn{2}{c}{} & & & & \multicolumn{1}{l|}{} & \multicolumn{9}{c|}{\textbf{Action Recognition} (Top-1 Accuracy)} & \multicolumn{9}{c}{\textbf{Text-to-video Retrieval} (Recall@1)} \\
\multicolumn{3}{c}{\multirow{-2}{*}{\textbf{Model}}} & \rotatebox{90}{\textbf{\#Parameters}} & \rotatebox{90}{\textbf{\#Samples Seen}} & \rotatebox{90}{\textbf{\#Frames}} & \rotatebox{90}{\textbf{Average}} & \rotatebox{90}{SSv2} & \rotatebox{90}{EK100} & \rotatebox{90}{EgoExo4D} & \rotatebox{90}{Kinetics-400} & \rotatebox{90}{COIN (SR)} & \rotatebox{90}{COIN (TR)} & \rotatebox{90}{CrossTask (SR)} & \rotatebox{90}{CrossTask (TR)} & \rotatebox{90}{\textbf{Average}} & \rotatebox{90}{MSR-VTT} & \rotatebox{90}{ActivityNet} & \rotatebox{90}{DiDeMo} & \rotatebox{90}{MSVD} & \rotatebox{90}{YouCook2} & \rotatebox{90}{PVD-Bench} & \rotatebox{90}{Dream-1k} & \rotatebox{90}{VDC-1k} \\ \toprule
 & & RN50 (224px) & 75M & & & 21.8 & 2.1 & 1.5 & 2.1 & 41.4 & 8.6 & 39.0 & 10.9 & 68.7 & 28.3 & 28.7 & 17.7 & 24.7 & 29.7 & 5.1 & 27.6 & 47.2 & 46.0 \\
 & & ViT-B (224px) & 124M & &  & 25.4 & 3.1 & 1.3 & 2.8 & 49.5 & 11.2 & 47.3 & 16.2 & 71.5 & 29.3 & 31.0 & 19.5 & 25.7 & 34.0 & 6.1 & 27.0 & 48.5 & 42.9 \\
\multicolumn{2}{c}{\multirow{-3}{*}{\textbf{CLIP}}} & ViT-L (336px) & 389M & \multirow{-3}{*}{12.8B} & \multirow{-3}{*}{8} & 30.7 & 3.8 & 3.7 & 2.6 & 58.3 & 14.7 & 63.5 & 20.8 & 78.5 & 35.3 & 35.9 & 23.4 & 30.7 & 41.9 & 7.9 & 36.7 & 56.8 & 49.3 \\ \midrule
 & & ViT-B (224px) & 375M & & & 33.9 & 5.2 & 2.3 & 4.5 & 57.8 & 20.6 & 69.9 & 27.7 & 82.9 & 39.6 & 40.2 & 25.0 & 32.1 & 48.6 & 13.8 & 52.1 & 60.9 & 43.7 \\
 & & ViT-L (384px) & 882M & & & 38.6 & 5.9 & 4.5 & 6.4 & 63.6 & 24.2 & 78.5 & 35.1 & 90.8 & 45.4 & 41.6 & 32.7 & 35.1 & 53.5 & 19.0 & 59.2 & 71.6 & 50.9 \\
\multicolumn{2}{c}{\multirow{-3}{*}{\textbf{SigLIP2}}} & ViT-g (384px) & 1.9B & \multirow{-3}{*}{40B} & \multirow{-3}{*}{8} & 39.8 & 6.1 & 6.1 & 5.6 & 68.0 & 26.0 & 80.4 & 35.1 & 90.8 & 47.5 & 43.4 & 33.9 & 38.9 & 56.0 & 22.2 & 60.4 & 73.0 & 52.5 \\ \midrule
 & & ViT-B (224px) & 448M & 58B &  & 37.2 & 5.8 & 3.3 & 6.0 & 65.4 & 21.5 & 77.1 & 26.9 & 91.8 & 44.9 & 46.5 & 35.4 & 35.3 & 49.1 & 15.2 & 59.8 & 68.7 & 49.2 \\
 & & ViT-L (336px) & 671M & 58B &  & 42.9 & 9.3 & 6.0 & 11.6 & 73.4 & 27.1 & 83.3 & 37.5 & 95.3 & 50.2 & 48.9 & 41.7 & 40.8 & 56.2 & 22.5 & 64.7 & 75.9 & 51.0 \\
\multicolumn{2}{c}{\multirow{-3}{*}{\textbf{PE-Core}}} & ViT-G (448px) & 2.3B & 86B & \multirow{-3}{*}{8} & 44.7 & 9.0 & 6.4 & 13.6 & \underline{\textbf{76.4}} & 29.0 & \underline{\textbf{86.0}} & 40.3 & \underline{\textbf{97.2}} & 58.1 & \underline{\textbf{51.6}} & 49.1 & 44.5 & \underline{\textbf{58.7}} & 26.0 & 77.0 & 89.2 & 68.5 \\ \midrule \midrule
\rowcolor[HTML]{EBF3FF} 
  & Stage 1 & & & 2.4B & 8 & 21.5 & 3.9 & 1.0 & 3.2 & 39.5 & 12.1 & 40.6 & 20.9 & 50.9 & 32.6 & 27.7 & 20.9 & 26.6 & 34.2 & 3.4 & 51.6 & 52.9 & 43.8 \\
\rowcolor[HTML]{EBF3FF} 
  & Stage 2 & & & 3.1B & 8 & 48.0 & 18.4 & 16.8 & 25.7 & 61.2 & 43.3 & 72.7 & 62.0 & 83.9 & 59.5 & 37.1 & 54.2 & 46.0 & 45.7 & 33.9 & 80.2 & 91.5 & 87.3 \\
\rowcolor[HTML]{EBF3FF} 
\multirow{-3}{*}{\textbf{VL-JEPA}} & Stage 3 & \multirow{-3}{*}{ViT-L (256px)} & \multirow{-3}{*}{1.6B} & 3.3B & 32 & \underline{\textbf{52.5}} & \underline{\textbf{19.3}} & \underline{\textbf{21.8}} & \underline{\textbf{33.2}} & 64.8 & \underline{\textbf{47.4}} & 79.4 & \underline{\textbf{64.5}} & 89.6 & \underline{\textbf{63.7}} & 40.0 & \underline{\textbf{64.9}} & \underline{\textbf{50.0}} & 49.0 & \underline{\textbf{40.4}} & \underline{\textbf{83.1}} & \underline{\textbf{93.3}} & \underline{\textbf{88.8}} \\
\bottomrule 
\end{tabular}%
}
\end{table*}

As shown in Tab.~\ref{tab:cls_ret}, with \textsc{Action100M}, \texttt{VL-JEPA} achieves higher average action recognition performance and average video retrieval performance. Thanks to strong \texttt{V-JEPA 2} encoder and dense fine-grained action annotation in \textsc{Action100M}, \texttt{VL-JEPA} is particularly strong on \textbf{motion-focused} tasks, such as Something-something-v2, EPIC-KITCHENS-100, EgoExo4D Keysteps, step recognition on COIN and CrossTask.

\begin{figure}
    \centering
    \includegraphics[width=1\linewidth]{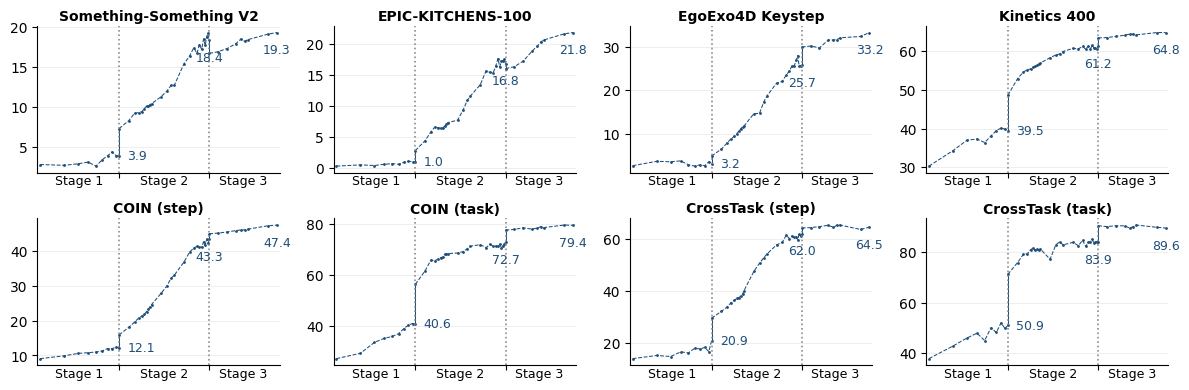}
    \caption{\textbf{Zero-shot action recognition accuracy in each training stage.} 
        \textbf{Stage 1}: image pretraining with single frame input; 
        \textbf{Stage 2}: main Action100M pretraining with 8 frames input; 
        \textbf{Stage 2}: 32 frames input with unfrozen encoder. 
    The x-axis of each stage represents number of training samples seen (\textit{i.e.,} number of iterations) in log-scale. The performance at the end of each stage is annotated.}
    \label{fig:performance_curve_classification}
\end{figure}

Fig.~\ref{fig:performance_curve_classification} visualize evolution of per-dataset accuracy in each stages, with x-axis being log-scale number of samples seen in each stage. We see that scaling yields consistent improvement, especially on motion-focused datasets mentioned earlier. We also see a significant jump from stage 1 to stage 2, indicating that image-only training alone is insufficient for action recognition.

\subsection{Effectiveness of \textsc{Action100M} Pipeline}

\begin{figure}[ht]
    \centering
    \includegraphics[width=0.95\linewidth]{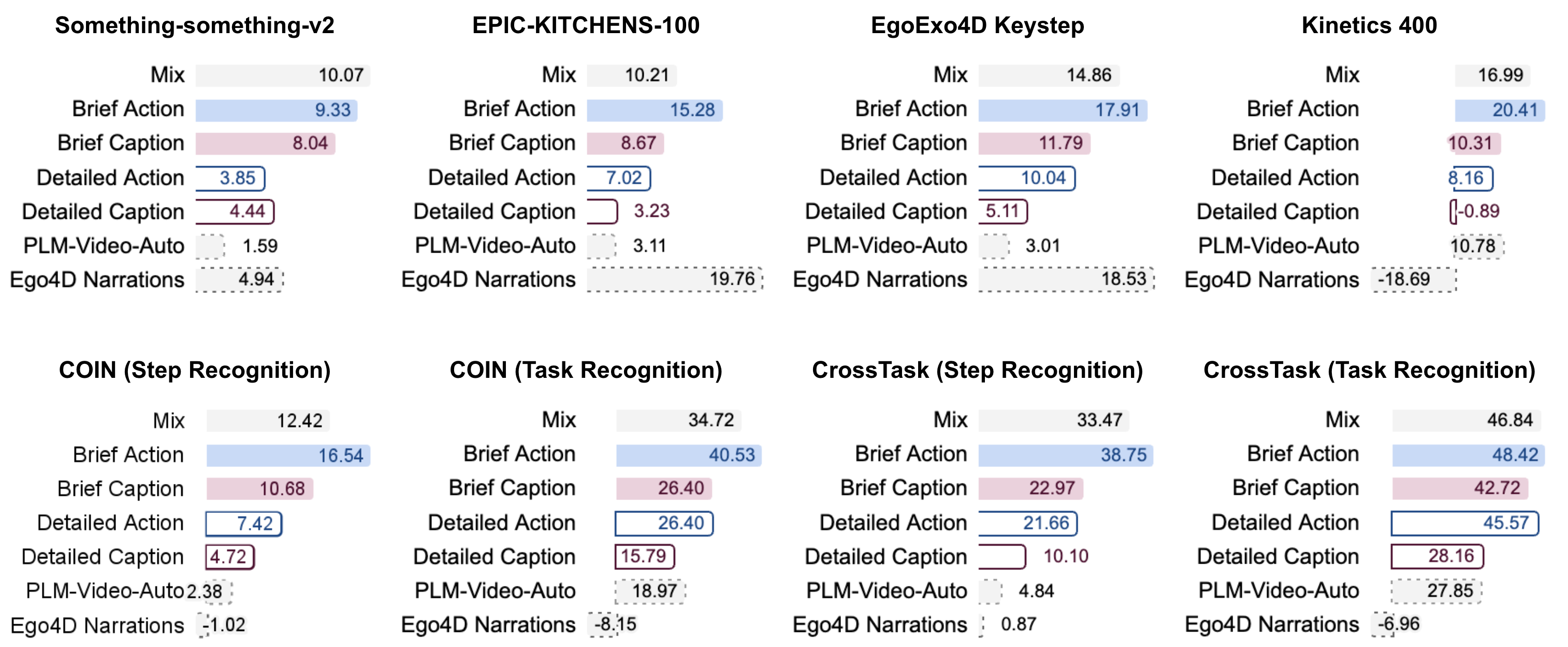}
    \caption{Performance improvements of stage 2 video pretraining with different data upon stage 1.}
    \label{fig:ablation_annotation_type_classification}
\end{figure}

We compare the effectiveness of different data under a controlled training setting. All models are initialized from the same stage 1 checkpoint, and further trained for 20k steps (20.48M samples seen). Fig.~\ref{fig:ablation_annotation_type_classification} visualizes the resulting performance improvement compared to the stage 1 initialization.

We see that brief action descriptions in \textsc{Action100M} are highly effective in terms of improving zero-shot action recognition. It outperform the direct PLM-3B pseudo labeling baseline, showing the effectiveness of hierachical captioning and LLM aggregation. Detailed captions in \textsc{Action100M} also show advantages over PLM-Video-Auto captions on most benchmarks. At the same time, \textsc{Action100M} is much larger than it (100M vs 3M). Training with Ego4D atomic action description significantly improve egocentric action recognition performance on EK-100 and EgoExo4D, while not being effective on other domains.

\subsection{Effectiveness of Semantic Resampling} 
\label{sec:semantic resampling}

A major challenge in large-scale action datasets is the long-tailed distribution of action frequencies, which can bias model training. To address this, we explore semantic resampling inspired by DINOv2~\citep{vo2024automatic} to promote a more balanced sampling of actions.

We begin by embedding all brief action descriptions from the \textsc{Tree-of-Captions} using \texttt{EmbeddingGemma-300M}~\citep{vera2025embeddinggemma}. To address redundancy, we deduplicate these actions by hashing their text, ensuring that repeated descriptions are consolidated. Following deduplication, we apply k-means clustering to the resulting embeddings, with cluster counts of $k=\{10^3,10^4,10^5\}$. This clustering groups semantically similar actions together, allowing us to control the granularity of the action space. From each cluster, actions are sampled uniformly with replacement until the target dataset size is reached, which ensures that both frequent and rare actions are adequately represented in the training data.

\begin{wrapfigure}{r}{0.38\linewidth}
    \centering
    \vspace{-15pt}
    \includegraphics[width=\linewidth]{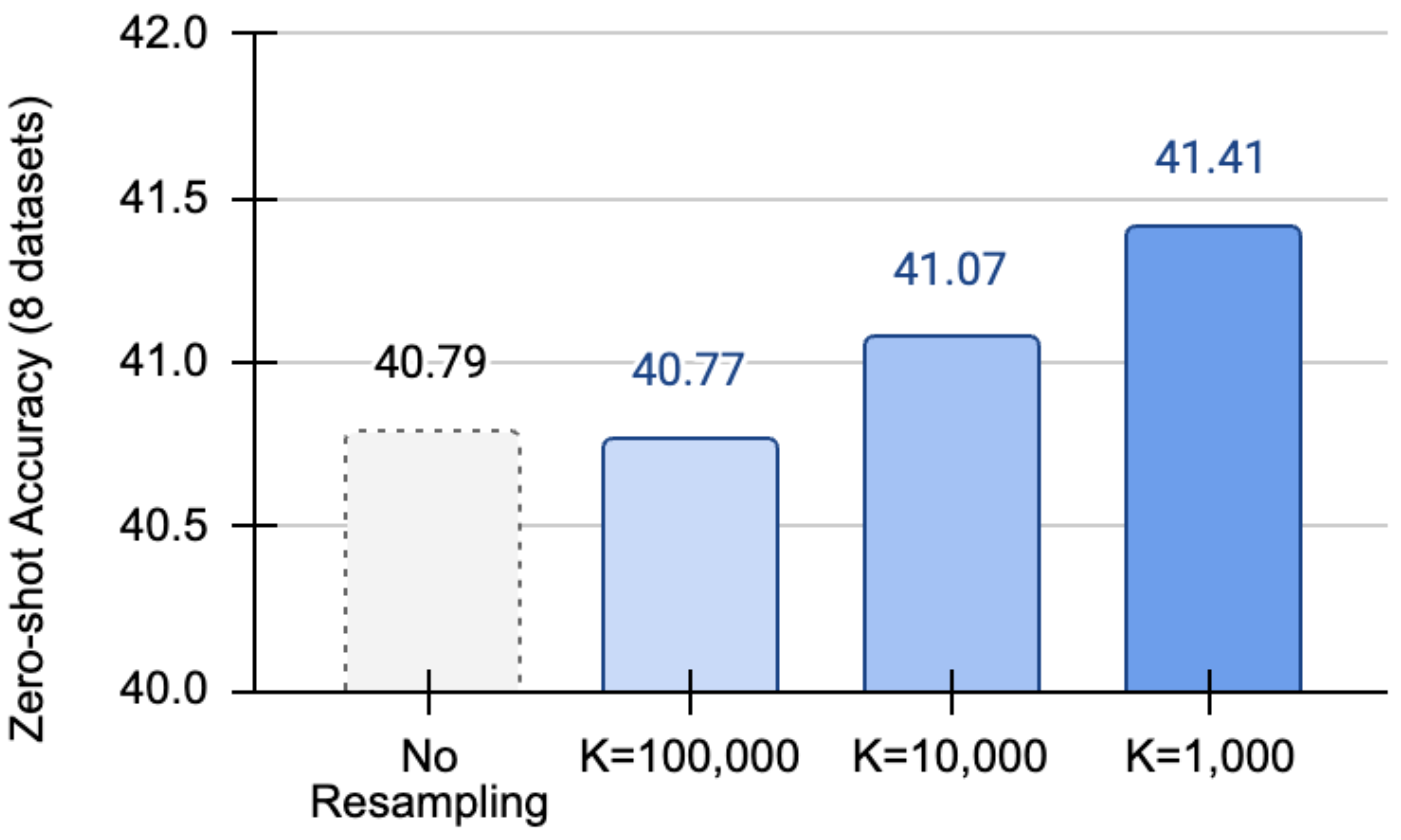}
    \vspace{-15pt}
    \caption{ \textbf{Effectiveness of semantic resampling.}
    }
    \vspace{-15pt}
    \label{fig:resampling}
\end{wrapfigure}

To empirically validate the effectiveness of this pipeline, we curated a 10M-action dataset using semantic resampling across all tested values of $k$. All models were initialized from the same stage 1 checkpoint and subsequently trained for 10k steps, corresponding to 10.24M samples seen. We report the average across the 8 benchmark zero-shot action classification datasets. As shown in Fig.~\ref{fig:resampling}, resampling with a smaller number of clusters leads to improved performance compared to training without resampling, highlighting the benefits of down-sampling frequent actions and up-sampling rare ones.

\section{Conclusion}

We introduced \textsc{Action100M}, a large-scale, open-domain dataset for action-centric video understanding built from 1.2M procedural videos and annotated into 147M temporally localized segments. Empirically, pretraining \texttt{VL-JEPA} on \textsc{Action100M} yields consistent scaling behavior and strong zero-shot transfer across a diverse set of action recognition and text-to-video retrieval benchmarks, with notable strengths on motion-focused and step-centric datasets. Controlled ablations further show the value of LLM-aggregated brief actions and hierarchical evidence over direct pseudo-labeling, and semantic resampling improves sample efficiency by mitigating long-tail redundancy. Overall, \textsc{Action100M} provides a practical route to scale open-vocabulary action understanding, and it enables future work on action anticipation, action-conditioned world models, and long-horizon planning that requires reasoning over multi-scale procedural structure.

\bibliographystyle{assets/plainnat}
\bibliography{paper}

\newpage
\appendix

\newpage
\section{Implementation Details}

We process each node in the \textsc{Tree-of-Captions} independently to generate structured action annotations. Nodes shorter than four seconds are discarded. For the remaining nodes, we query a large reasoning model (\texttt{GPT-OSS-120B}\footnote{\url{https://huggingface.co/openai/gpt-oss-120b}}) to infer clean, structured textual representations that unify and denoise information from multiple caption sources.

For each node, we construct an input prompt that combines: formatted \textsc{Tree-of-Captions} of the current node and the global root node, and video metadata context-including title, description, ASR transcript. \textsc{Tree-of-Captions} are formatted as a depth-first traversal into a Markdown-style text stream. The serialized input is then processed by \texttt{GPT-OSS-120B} under a 3 rounds of \textsc{Self-Refine} procedure. In the first round, the model performs structured reasoning with a \texttt{high} reasoning effort setting to produce an initial draft of the action summary. In subsequent rounds, the same model revisits the previous output together with the original context, iteratively correcting factual errors, resolving inconsistencies, and removing unsupported statements. The prompt to the \texttt{GPT-OSS-120B} is as follows:
\begin{lstlisting}[language={},basicstyle=\ttfamily\small]
# Video metadata

{video_metadata}

# Global video context

{formatted_global_tree_of_captions}

# Current segment to be processed

{formatted_current_tree_of_captions}

# Your Task

The task is to extract structured information from a video segment. The segment spans from {start_time} to {end_time} seconds, within a larger video that runs from {global_start_time} to {global_end_time} seconds. You will be provided with hierarchical captions generated by models operating on local windows or frames. These captions may contain errors or hallucinations. Your goal is to carefully aggregate the information from the captions to produce an accurate, coherent description of this specific segment.

Guidelines and requirements:

- Focus on what is visually observable, emphasizing both 1) physical motion, procedural actions, and 2) appearance information, background or text if possible.
- The timestamps in video metadata and markdown titles are generally reliable, but those inside the captions are not.  
- Use global captions and video metadata only for disambiguation. Do not add visually unobservable information (e.g., content of speech, names that cannot be inferred from the local video segment) to the results.
- The result covers the current segment ({start_time} to {end_time}) only, not the entire video.
- Ignore captions from very short edges at the start or end of the segment if they are semantically discontinuous (e.g., due to scene transitions). Focus on the main central portion of the segment.
- Due to the limited perception of local models and possible hallucinations, there may be inconsistencies among captions. 
- Be cautious and conservative, and rely on the majority consensus.  
- Think hard. Perform in-depth reasoning to discuss the provided captions and global context, then output a JSON containing final results in plain English text.
- For all fields, use coherent full sentences with proper capitalization and punctuation. Use concise noun phrases without unnecessary qualifiers or parenthetical clarifications.


### Task 1. Summarization

Generate both a short informative caption and a comprehensive, dense summary of the video segment. Describe events in chronological order, but do not mention any exact timestamps in the summary. Include 

### Task 2. Action Identification

Identify the main actor and the physical action performed in the current segment. Provide both a brief description that represents the overall action step, and a detailed description that contains sufficient procedural detail. Use "N/A" (without further explaination) if there are no visible actors or physical actions (e.g., static).


# Response Formats
## output
\end{lstlisting}

\begin{lstlisting}[language={},basicstyle=\ttfamily\small]
{
    "type": "object", "properties": {
        "summary": {"type": "object", "properties": {
            "brief": {
                "type": "string", 
                "description": "Single sentence video caption."
                },
            "detailed": {
                "type": "string",
                "description": "Detailed, comprehensive description."
                },
            },
        },
        "action":  {"type": "object",  "properties": {
            "brief": {
                "type": "string",
                "description": "A single verb phrase (no -ing forms) brifly summarizing the overall action content."
                },
            "detailed": {
                "type": "string",
                "description": "A single imperitive sentence describing how the action is performed with more details."
                },
            "actor": {
                "type": "string", 
                "description": "Single sentece or an imformative noun phrase describing who is performing the action."
                },
            },
        },
    },
    "required": ["summary", "action"]
}
\end{lstlisting}

For \textsc{Self-Refine}, we append the following instruction in addition:

\begin{lstlisting}[language={},basicstyle=\ttfamily\small]
Now, carefully analyze, verify, and revise the previous draft so that it is fully accurate, faithful to the provided content, and strictly adheres to all stated guidelines and requirements.
\end{lstlisting}

\newpage
\section{Statistics of Duplications and Semantic Resampling}

\begin{figure*}[h]
    \centering
    \includegraphics[width=0.65\linewidth]{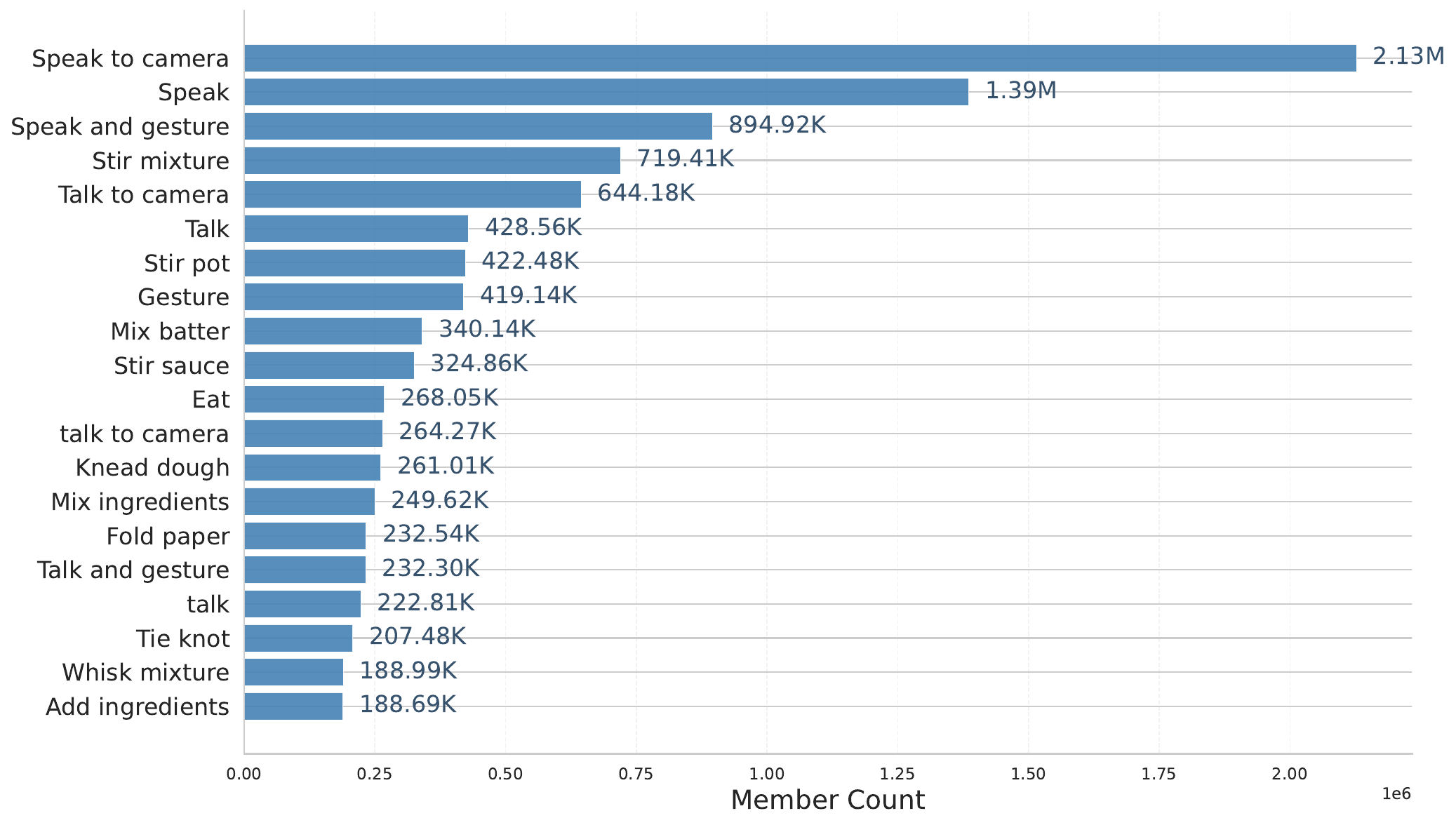}
    \caption{Distribution of duplicated brief action descriptions.}
    \label{fig:exact_duplicates}
\end{figure*}

The scale of redundancy in the raw data is substantial. A significant portion of brief action descriptions are repeated, reflecting the inherent long-tailed and redundant nature of large-scale video action data. During the de-duplication step, we identify 7.58 million duplicate groups, which together account for 141.8 million duplicate instances. The remaining action texts are unique, each occurring only once in the dataset. The distribution of these exact duplicates is visualized in Figure~\ref{fig:exact_duplicates}.

\begin{figure*}[t]
  \centering

  \begin{subfigure}{\textwidth}
    \centering
      \includegraphics[width=\linewidth]{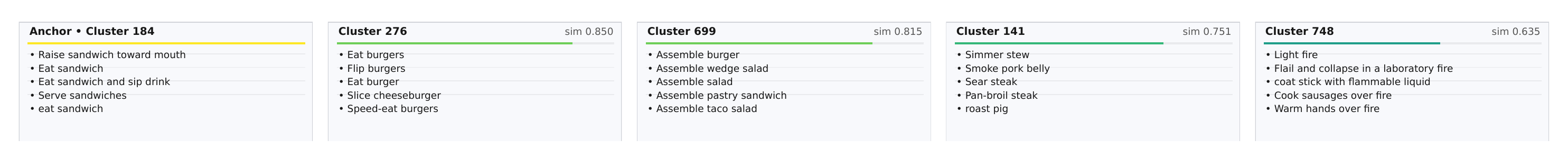}
      \includegraphics[width=\linewidth]{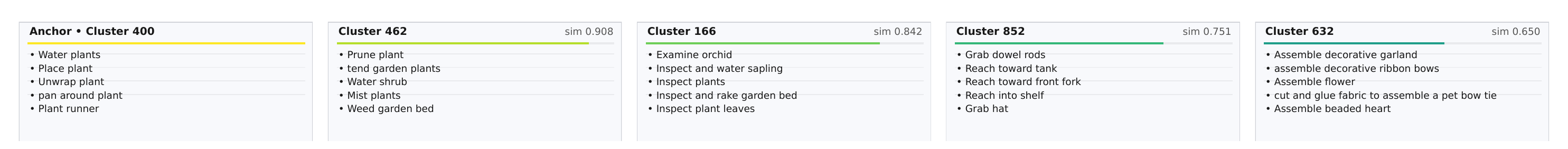}
      \subcaption{Distribution of texts within clusters for $k=10^3$}
      \label{fig:subfig1b}
  \end{subfigure}

  \vspace{0.7em} 

  \begin{subfigure}{\textwidth}
    \centering
      \includegraphics[width=\linewidth]{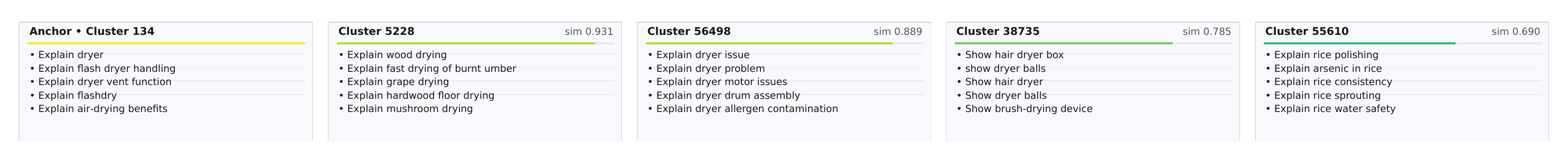}
      \includegraphics[width=\linewidth]{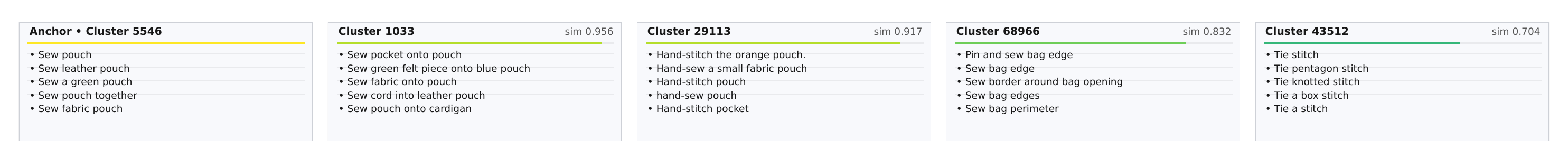}
      \subcaption{Distribution of texts within clusters for $k=10^4$}
      \label{fig:subfig2b}

  \end{subfigure}

  \vspace{0.7em}

  \begin{subfigure}{\textwidth}
    \centering
      \includegraphics[width=\linewidth]{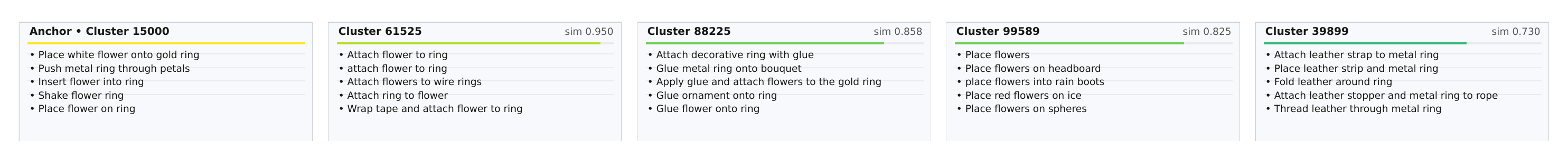}
      \includegraphics[width=\linewidth]{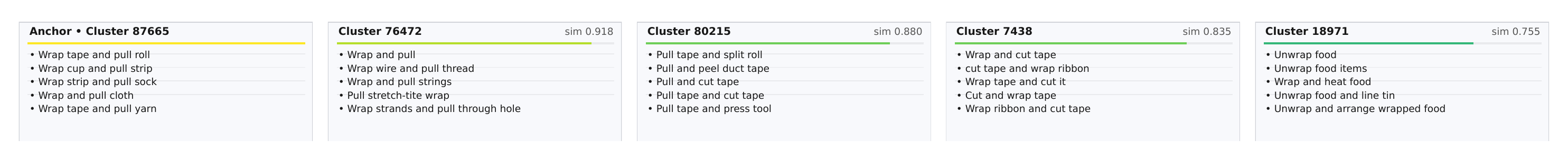}
      \subcaption{Distribution of texts within clusters for $k=10^5$}
      \label{fig:subfig3}
  \end{subfigure}

  \caption{\textbf{Understanding cluster distribution after the sementic clustering (deduplication+kmeans):} Examples of selected anchor clusters (left) and their closest neighboring actions (in decreasing order of similarity, right) for different values of $k$, illustrating the effectiveness of the clustering approach. We show 5 random texts within each cluster.  \textit{sim} of the neighboring clusters denote the cosine similarity to the anchor cluster. Clusters consistently group together actions with similar semantics across all $k$ values.  Lower $k$ values ($10^3$) yield broader, more diverse clusters, while higher $K$ values ($10^5$) produce highly specific clusters. }
  \label{fig:sample-texts}
\end{figure*}

After deduplication, we apply k-means clustering to the action texts, grouping them into $k=\{10^3, 10^4, 10^5\}$ clusters. Each cluster represents a set of semantically similar actions, enabling us to control the granularity of the action space. Figure~\ref{fig:sample-texts} illustrates the effectiveness of this clustering: for each $k$, we present examples of selected anchor clusters and their closest neighbors (ranked by cosine similarity), with five random action texts shown per cluster. Across all settings, the clustering pipeline consistently groups together actions with similar semantics, as evidenced by the high cosine similarity between neighboring clusters.

\begin{figure*}
    \centering
    \includegraphics[width=0.95\linewidth]{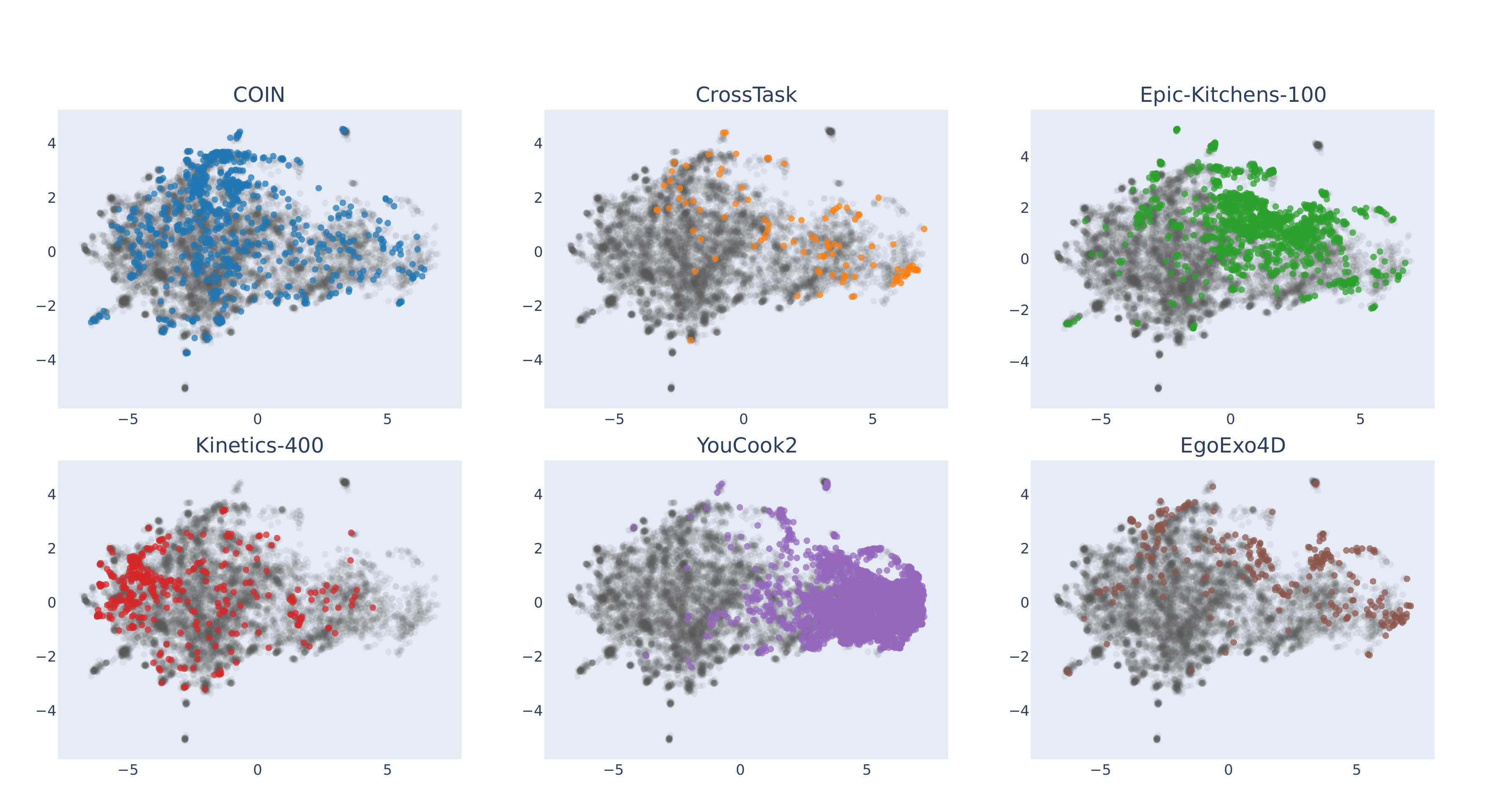}
    \caption{\textbf{UMAP visualization of the overlap between the semantic clusters and downstream datasets.} Each panel containing samples from a specific downstream dataset (colored points) and their overlap with the $k=10^4$ cluster of the \textsc{Action100M} dataset.  This highlights the diversity and coverage of the \textsc{Action100M} dataset with respect to multiple downstream benchmarks.}
    \label{fig:overlap-downstream}
\end{figure*}

To assess the coverage and diversity of our clustered action space, we analyze the overlap between the $k=10^4$ clusters and several standard downstream action recognition datasets. Figure~\ref{fig:overlap-downstream} presents a UMAP visualization, where each panel highlights the clusters containing samples from a specific benchmark (COIN, CrossTask, Epic-Kitchens-100, Kinetics-400, YouCook2, EgoExo4D). The varying distributions of colored points across the panels demonstrate that the \textsc{Action100M} clusters provide broad and diverse coverage of the action space, with different downstream datasets mapping to distinct but overlapping regions. This analysis confirms that our semantic clustering approach not only reduces redundancy and enhances diversity, but also ensures strong representational alignment with a wide range of downstream tasks.

\end{document}